广义去噪扩散编码模型（gDDCM）：使用预训练的扩散模型对图像进行分词

# 广义去噪扩散编码模型（gDDCM）：使用预训练的扩散模型对图像进行分词

孔飞，石小爽*

(电子科技大学 计算机科学与工程学院，成都 611731)

**摘要**：去噪扩散模型已成为图像生成领域的主流范式，而将连续的图像数据离散化为令牌（Token）是图像与Transformer等架构高效结合的关键步骤。尽管去噪扩散编码模型（DDCM）率先利用预训练扩散模型进行图像分词，但其严格依赖于传统的离散时间DDPM架构，难以适配当前先进的连续时间模型（如流匹配模型、一致性模型等），且在高噪区域存在采样效率低下的问题。为此，本文提出了广义去噪扩散编码模型（Generalized Denoising Diffusion Coding Model, gDDCM）。我们构建了一个统一的理论框架，提出了一种通用的“去噪-回溯”采样策略，通过结合确定性的常微分方程（ODE）去噪步与基于残差对齐的噪声注入步，成功解决了DDCM在主流扩散模型的适配问题。此外，我们引入了回溯参数$p$，显著提升分词的稳定性与质量。在CIFAR10和LSUN Bedroom数据集上的广泛实验表明，gDDCM不仅实现了对主流扩散模型变体的全面兼容，其在重构质量和感知保真度上均优于基线方法DDCM。

# Generalized Denoising Diffusion Codebook Models (gDDCM): Tokenizing images using a pre-trained diffusion model

KONG Fei, SHI Xiaoshuang*

(School of Computer Science and Engineering, University of Electronic Science and Technology of China, Chengdu 611731, China)

**Abstract** Denoising diffusion models have emerged as a dominant paradigm in image generation. Discretizing image data into tokens is a critical step for effectively integrating images with Transformer and other architectures. Although the Denoising Diffusion Codebook Models (DDCM) pioneered the use of pre-trained diffusion models for image tokenization, it strictly relies on the traditional discrete-time DDPM architecture. Consequently, it fails to adapt to modern continuous-time variants—such as Flow Matching and Consistency Models—and suffers from inefficient sampling in high-noise regions. To address these limitations, this paper proposes the Generalized Denoising Diffusion Codebook Models (gDDCM). We establish a unified theoretical framework and introduce a generic "De-noise and Back-trace" sampling strategy. By integrating a deterministic ODE denoising step with a residual-aligned noise injection step, our method resolves the challenge of adaptation. Furthermore, we introduce a backtracking parameter $p$ and significantly enhance tokenization ability. Extensive experiments on CIFAR10 and LSUN Bedroom datasets demonstrate that gDDCM achieves comprehensive compatibility with mainstream diffusion variants and significantly outperforms DDCM in terms of reconstruction quality and perceptual fidelity.

**Key words** Diffusion Models; Image Tokenization; gDDCM; Consistency Models; Flow Matching Models

## 1 介绍

扩散模型（Diffusion Models）凭借其生成高质量、多样化图像的能力，已在图像生成领域取得显著成就，并展现出零样本修复（Zero-shot Inpainting）和音频生成等广泛的应用潜力。然而，高昂的计算成本和缓慢的采样速度构成了其主要瓶颈。传统的去噪扩散概率模型（DDPM [1]）需要通过迭代去噪过程，逐步将高斯噪声转化为目标数据分布。

为解决这一问题，研究界提出了多种改进方案。分数匹配模型（Score-based Models [2]）将DDPM扩展至连续时间域，证明了扩散过程可由随机微分方程（SDE）描述。值得注意的是，该框架揭示了相同的边缘分布可以对应不同的前向加噪方差；当方差为零时，采样过程退化为常微分方程（ODE），从而

孔飞，kong13661@outlook.com

*通信作者E-mail：xsshi2013@gmail.com

允许利用高效ODE求解器加速采样。在此基础上，一致性模型（Consistency Models [3]）通过蒸馏技术，使得采样路径上任意一点直接映射至终点，实现了单步生成的突破。此外，流匹配模型（Flow Matching [4]）及重整流模型（Rectified Flow）通过重新定义速度场和拉直采样轨迹（Straightening Trajectories），进一步提升了采样效率。尽管上述变体采用了不同的采样路径设计，但它们在理论上共享相似的边缘分布特性。

图像离散化[5][6]在人工智能之中有着广泛的应用，特别是随着Transformer架构[7]已成为生成式建模的主流范式，大语言模型（LLMs）中，基于“预测下一个令牌（Next-token Prediction）”的机制已展现出卓越的性能。为了将这一范式迁移至图像生成，通常需要通过“分词（Tokenization）”将连续的图像数据离散化。主流方法如VQVAE [8]和VQGAN [9]采用量化重构的方式将图像编码为2D令牌网格；而TiTok [10]则利用QFormer [11]生成更适配自回归生成的1D单向依赖令牌。

然而，上述方法通常需要训练专门的分词器模型。据我们所知，去噪扩散编码模型（DDCM [12]）是首个利用预训练扩散模型进行图像离散化和压缩的工作。DDCM通过在逆向去噪过程中引入特定的偏置噪声，使其向重构样本逼近，从而实现编码。尽管DDCM具有开创性，通过我们的分析，发现其仍存在显著局限性：1. 适用性受限：DDCM严格基于DDPM构建。然而，如前所述，扩散模型领域已涌现出流匹配、一致性模型等多种高效变体，DDCM无法直接适配这些先进架构；2. 采样效率低： 在高噪声水平的初始阶段，DDCM往往需要过多的迭代步数，导致计算资源的浪费。3. 原论文主要关注图像压缩，缺乏对分词性能的深入分析。

为了解决上述问题，本文提出了广义深度去噪编码模型（Generalized Denoising Diffusion Coding Model, gDDCM）。本文的主要贡献总结如下：

- 提出了统一的理论框架：我们建立了一个通用的理论形式，不仅证明了其数学合理性，还成功将DDCM的思想扩展至主流的扩散模型变体（包括一致性模型，分数匹配模型及流匹配模型），打破了原有方法的架构限制。
- 实现了高效且适配LLM的分词机制：gDDCM生成的令牌天然具备1D结构和单向依赖性，且易于扩展至2D形式。这种特性使其与大语言模型的“Next-token Prediction”模式具有天然的适配性，无需额外的复杂适配层。
- 验证了优越的性能表现：我们在CIFAR10，以及LSUN Bedroom数据集上进行了广泛的实验。结果表明，gDDCM不仅能够适配各种主流扩散模型变体进行有效分词，而且在重构质量和效率上均优于原始DDCM。

## 2 相关工作

### 2.1 扩散模型及其变体

扩散模型已成为图像生成领域的主流范式。 与依赖生成器和判别器对抗博弈的传统生成对抗网络（GANs[13]）不同，扩散模型通过逆向模拟高斯噪声的扩散过程来重建数据样本。扩散模型在训练稳定性及模式覆盖方面表现出显著优越性。

扩散模型的研究已从早期的离散时间步长的去噪扩散模型[1]演变为基于分数[2]的连续时间生成模型。 现有研究证明，扩散过程可以由随机微分方程描述，且存在一个对应的概率流常微分方程，二者具有相同的边缘分布但对应不同的加噪轨迹。基于此理论框架，一致性模型[3]通过对ODE轨迹进行蒸馏，实现了单步生成的高效采样。

此外，流匹配模型[4]作为一种更通用的框架被提出。 虽然流匹配与分数匹配模型在边缘分布上保持一致，但其构建了不同的概率流向量场，从而产生了差异化的加噪与去噪轨迹。在流匹配的基础上，重整流模型进一步引入了“重整（Reflow）”操作，通过拉直常微分方程的轨迹线，有效降低了传输代价，最终赋予模型单步采样的能力。

扩散模型的多种变体主要在于加噪和降噪的轨迹不同。然而他们的边缘分布（在某个时刻$t$的数据的分布）是相似的。通常为$p_t(\boldsymbol{x}_t) = f(t)\boldsymbol{x} + g(t)\boldsymbol{\epsilon}$，$\epsilon \sim \mathcal{N}(0,1)$，其中$\boldsymbol{x}$的分布是数据集分布。下面几节详细介绍这几种模型。

#### 2.1.1 去噪扩散概率模型（DDPM）

扩散模型通过对从高斯噪声开始的正向扩散过程的逆过程进行建模来生成样本。正向扩散过程可以视为添加噪声的过程，定义如下：

$$\boldsymbol{x}_t = \sqrt{\alpha_t}\boldsymbol{x}_{t-1} + \sqrt{\beta_t}\boldsymbol{\epsilon}_t,\ \boldsymbol{\epsilon}_t \sim \mathcal{N}(0, I). \quad (1)$$

随着$t$增大，$\beta_t$逐渐增大，最终$\boldsymbol{x}_t$越来越近似于随机高斯噪声。采样的时候，从标准高斯分布中采样一个噪声，然后使用逆向扩散过程进行去噪。记$\{\boldsymbol{x}'_t\}$为反向扩散过程序列，可以证明$\boldsymbol{x}'_t$依然服从高斯分布。假设其方差与前向扩散过程相同，$\boldsymbol{x}'_t$的均值定义为：

$$\tilde{\mu}_t = \frac{1}{\sqrt{a_t}}\left(\boldsymbol{x}_t - \frac{\beta_t}{\sqrt{1-\overline{\alpha}_t}}\overline{\epsilon}_\theta(\boldsymbol{x}_t, t)\right),$$

其中$\overline{\alpha}_t = \prod_{k=0}^{t} \alpha_k$并且$\overline{\alpha}_t + \overline{\beta}_t = 1$。反向扩散过程变为：

$$x_{t-1} = \tilde{\mu}_t + \sqrt{\beta_t}\boldsymbol{\epsilon}, \boldsymbol{\epsilon} \sim \mathcal{N}(0, I). \quad (2)$$

模型需要学习前向的噪音$\overline{\boldsymbol{\epsilon}}_t$。损失函数定义为：

$$\mathbb{E}_{\boldsymbol{x}_0, \overline{\boldsymbol{\epsilon}}_t}\left[\left\|\overline{\boldsymbol{\epsilon}}_t - \boldsymbol{\epsilon}_\theta\left(\sqrt{\overline{\alpha}_t}\boldsymbol{x}_0 + \sqrt{1-\overline{\alpha}_t}\overline{\boldsymbol{\epsilon}}_t, t\right)\right\|^2\right].$$

逆向过程可以看做去噪的过程。扩散模型逆向过程需要的步骤较多，如在CIFAR10上面，需要迭代总步数$T = 1000$。

2.1.2 分数匹配模型

基于分数的模型将离散时间扩散过程转化为连续时间过程，并采用随机微分方程（SDE）来表达扩散过程。此外，在保证边缘分布相同的情况下，前向和反向过程不再局限于扩散过程。

分数匹配模型采用的前向过程定义为：

$$d\boldsymbol{x} = \left(f_t(\boldsymbol{x}) - \frac{1}{2}(g_t^2 - \sigma_t^2)\nabla_x \log p_t(\boldsymbol{x})\right)dt + \sigma_t d\boldsymbol{w}. \quad (3)$$

其中$\boldsymbol{w}$是标准前向维纳过程。相应的反向过程为：

$$d\boldsymbol{x} = \left(f_t(\boldsymbol{x}) - \frac{1}{2}(g_t^2 + \sigma_t^2)\nabla_x \log p_t(\boldsymbol{x})\right)dt + \sigma_t d\overline{\boldsymbol{w}}, \quad (4)$$

其中$\overline{\boldsymbol{w}}$是逆向时间标准维纳过程。当$\sigma_t \equiv 0$时，前向和逆向过程变为常微分方程的形式，

$$d\boldsymbol{x} = \left(f_t(\boldsymbol{x}) - \frac{1}{2}g_t^2\nabla_x \log p_t(\boldsymbol{x})\right)dt. \quad (5)$$

当$g_t \equiv \sigma_t$时，则分数匹配模型等价于$T = \infty$的扩散模型，此时前向过程为：

$$d\boldsymbol{x} = f_t(\boldsymbol{x})dt + \sigma_t d\boldsymbol{w}. \quad (6)$$

在常微分方程的情况下，可以方便的使用对应的求解器。在实际应用中，现有的工作通常会选择$f_t(\boldsymbol{x}) = f(t)\boldsymbol{x}$。通过训练模型对$\nabla_{\boldsymbol{x}} \log p_t(\boldsymbol{x})$进行估计，可最终用来生成样本。可以证明，模型的损失函数跟DDPM相似。

2.1.3 一致性模型

扩散模型的采样时间长的问题。为了减少采样次数，一致性模型被提出来。如果一个函数在轨迹上的每个点输出都相同，则该函数被称为一致性函数。形式化地说，给定一条轨迹$\{x_t\}$其中$t \in [0, T]$，该函数满足$f(\boldsymbol{x}_{t_1}) = E[f(\boldsymbol{x}_{t_2})]$。如果该轨迹不是概率轨迹，则上述公式中的期望符号可以去掉，也即$f(\boldsymbol{x}_{t_1}) = f(\boldsymbol{x}_{t_2})$。[14]提出一致性扩散模型（CDM），其证明了当前向扩散过程满足 $dx_t = g(t)dw_t$时，$h(x, t) = \nabla \log q_t(x) g^2(t) + x$是一个一致性函数。他们在训练过程中额外添加了一致性正则化以提高模型的采样效率。[3]提出了一致性模型（CM）。与一致性扩散模型不同，一致性模型利用确定性采样，通过学习从轨迹上的任意点$x_t$到起点$\boldsymbol{x}_0$的映射，来获得一步采样的模型。当训练一个扩散模型来获取轨迹$\boldsymbol{x}_t$时，该方法称为一致性蒸馏。当使用条件轨迹来近似非条件轨迹时，该方法称为一致性训练。

2.1.4 流匹配模型

从公式(3)可以得出结论，分数匹配模型的路径与边缘分布的梯度$\nabla_{\boldsymbol{x}} \log p_t(\boldsymbol{x})$有关。与分数匹配模型不同，流匹配模型放松了这个限制。通过保留与分数匹配模型类似的边缘分布，流匹配模型提出了新的加噪和降噪路径。流匹配模型核心思想是学习两个概率分布之间由速度场描述的流动匹配关系。形式化地，给定数据分布$\boldsymbol{x} \sim p_{\text{data}}(\boldsymbol{x})$和另一分布（通常为容易采样的噪音分布）$\boldsymbol{\epsilon} \sim p(\boldsymbol{x})$，可以构建随时间$t$变化的条件流路径$\boldsymbol{z}_t = \alpha_t\boldsymbol{x} + \beta_t\boldsymbol{\epsilon}$。在使用流匹配模型的时候，通常会选择$\boldsymbol{\epsilon} \sim \mathcal{N}(0,1)$，$\alpha_t = 1 - t$以及$\beta_t = t$。条件速度被定义为$\boldsymbol{v}(\boldsymbol{z}_t, t|\boldsymbol{x}) = \alpha_t'\boldsymbol{x} + \beta_t'\boldsymbol{\epsilon}$。边缘速度（非条件速度）为$\boldsymbol{v}(\boldsymbol{z}_t, t) = \mathbb{E}_{\boldsymbol{x} \sim p_{data}(\boldsymbol{x})}[\boldsymbol{v}(\boldsymbol{z}_t, t|\boldsymbol{x})]$。给出边缘速度，可以通过常微分方程定义非条件流路径$d\boldsymbol{z}_t = \boldsymbol{v}(\boldsymbol{z}_t, t)dt$。为了生成图像，需要训练一个模型估计$\boldsymbol{v}(\boldsymbol{z}_t, t)$，这等价于通过下面损失函数训练模型：

$$\mathbb{E}_{t, \boldsymbol{x}, \boldsymbol{\epsilon}}\|\boldsymbol{v}_{\boldsymbol{\theta}}(\boldsymbol{z}_t, t) - \boldsymbol{v}(\boldsymbol{z}_t, t|\boldsymbol{x})\|^2. \quad (7)$$

这个损失函数和扩散模型是相似的。重整流模型可以通过训练好的流匹配模型重新生成数据计算条件分布。此时由于$\boldsymbol{x}$与$\boldsymbol{\epsilon}$不再是完全随机的对应关系，相比于完全随机的对应关系生成的边缘速度场，此时的边缘速度场里，从随机噪音转换到样本的路径也更接近线性。经过重新训练的模型就可以使用更少的采样生成样本。这称为重整流模型。

2.2 去噪扩散编码模型（DDCM）

最近有工作提出使用去噪扩散模型[12]对图像进行压缩或分词（tokenizing），将连续的图像转换为离散的令牌（token）。具体而言，他们首先通过从$\mathcal{N}(0,1)$中采样$K$个样本构成$\mathcal{E}_i$，其中每个去噪的采样时间$t$对应一个$i$。他们的实验展示了将公式(2)中的噪音$\boldsymbol{\epsilon}$转换为从$\mathcal{E}_i$采样，仍然能够生成高质量的图像。在这个基础上，他们提出在挑选噪音$\boldsymbol{\epsilon}$的时候，使用 $\underset{\epsilon \in \mathcal{E}_i}{\operatorname{argmax}}\left(\boldsymbol{x} - \boldsymbol{x}_{0|\boldsymbol{x}_i}\right) \cdot \boldsymbol{\epsilon}$近似$g_i^2\nabla_{\boldsymbol{x}_i} \log p_i(\boldsymbol{x}|\boldsymbol{x}_i)$。

他们提出的方法适用于扩散模型以及分数匹配模型

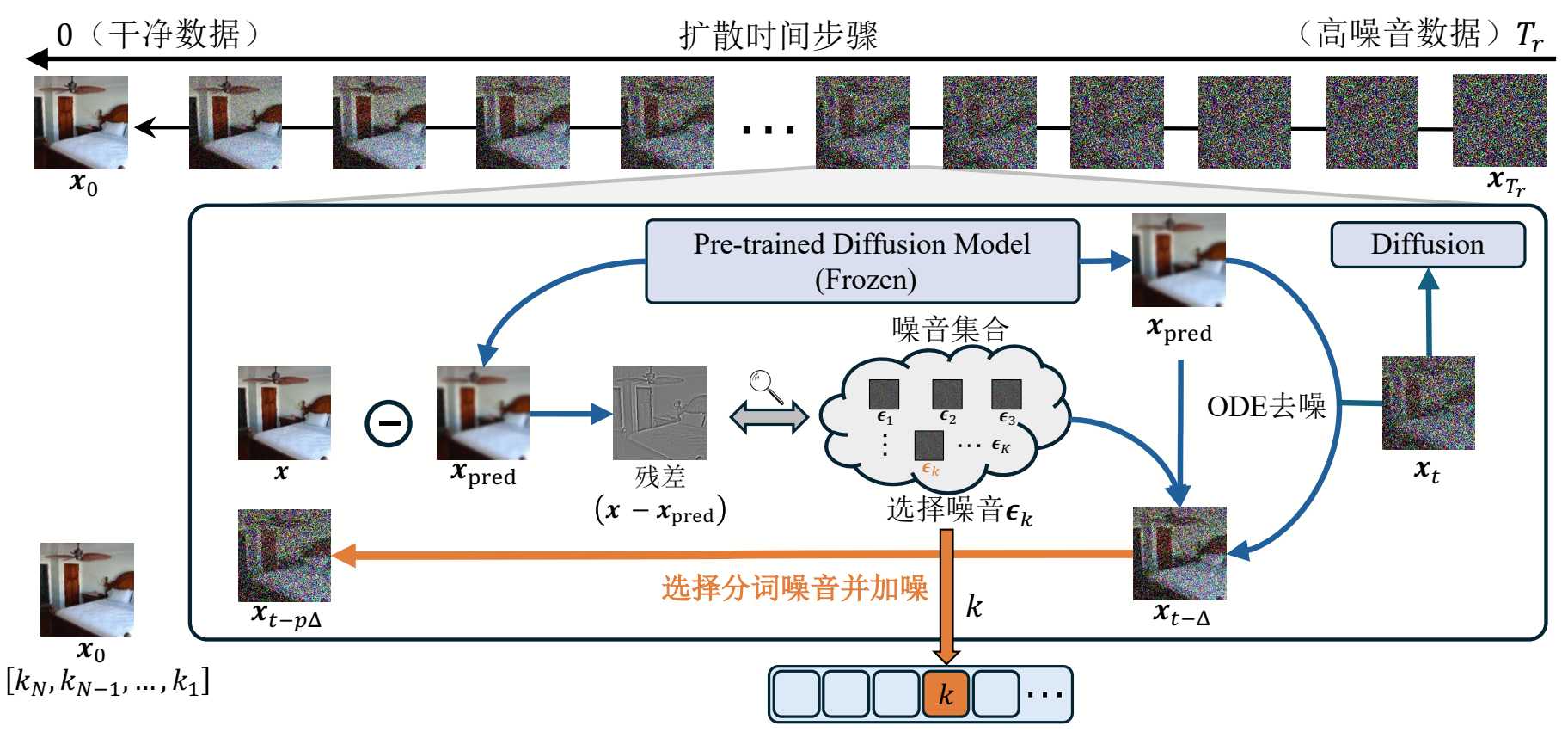

**图 1** 所提广义去噪扩散编码模型（gDDCM）的示意图。从某个时间$T_r$开始，经过初始化，输入进预训练的扩散模型。扩散模型的输出为$\boldsymbol{x}_{\mathrm{pred}}$。经过ODE去噪后获得$\boldsymbol{x}_{t-\Delta}$。此时计算需重构样本$\boldsymbol{x}$与$\boldsymbol{x}_{\mathrm{pred}}$的残差，从噪音集合中选择噪音，将索引放入到分词集合当中。$\boldsymbol{x}_{t-\Delta}$添加选择的噪音获得$\boldsymbol{x}_{t-p\Delta}$。迭代多次，直到生成$\boldsymbol{x}_0$。

(公式(4))在$g_t \equiv \sigma_t$时的情况。他们证明，在这种情况下，若有：

$$\|\boldsymbol{\epsilon} - g_i^2 \nabla_{\boldsymbol{x}_i} \log p_i\,(\boldsymbol{x}_0|\boldsymbol{x}_i)\| = 0,$$

则他们的方法等价于采样的时候使用：

$$\boldsymbol{x}_{i-1} = \boldsymbol{x}_i - f_{\mathrm{i}}(\boldsymbol{x}_i) + g_i^2 \nabla_{\boldsymbol{x}_i} \log p_i\,(\boldsymbol{x}_i|\boldsymbol{x}_0)$$

进行采样。

## 3　方法

正如在2.2节中所述，去噪扩散编码模型只适用于DDPM以及分数匹配模型的一个特例上。为了将DDCM扩展到扩散模型的变体上，我们提出了广义去噪扩散编码模型（gDDCM）。为了书写放标，在不引起歧义的情况下，我们使用$\boldsymbol{x}_0$表示$\boldsymbol{x}_{0|\boldsymbol{x}_t}$，使用$\boldsymbol{x}$表示从$p_{\mathrm{data}}$中采样的样本，使用$\boldsymbol{\epsilon}$表示模型的预测噪音或者从正态分布当中采样的随机噪音。

### 3.1　所提方法

#### 3.1.1　广义去噪扩散编码模型

尽管 DDCM 在基于DDPM的分词任务中表现出色，但将其扩展至其他扩散模型变体时存在障碍。首先，对于分数匹配模型，DDCM假设的$g_t \equiv \sigma_t$关系不再普遍成立，这导致噪声权重的变化会破坏原有的编码机制。其次，对于流匹配模型和一致性模型，其核心挑战在于采样过程的确定性。流匹配的逆向过程通常由常微分方程描述，而一致性模型则通过单步映射直接生成样本，无需通过求解微分方程进行迭代。这意味着在这些模型的采样路径中，缺乏一个显式的随机噪声项$\boldsymbol{\epsilon}$供DDCM进行操作（即无法通过替换噪声来嵌入离散化信息）。如图 2所示，直接将DDCM应用于流匹配模型会导致生成结果崩溃，无法得到有意义的样本。

为了克服上述障碍，我们需要挖掘不同扩散变体背后的共性。观察到尽管不同模型（如2.1所述）定义了不同的加噪与去噪概率流轨迹，但它们的边缘分布通常遵循相似的形式：

$$\boldsymbol{x}_t = s(t)\boldsymbol{x} + \Sigma(t)\boldsymbol{\epsilon}. \tag{8}$$

其中$\Sigma(t) = s(t)\sigma(t)$。尽管在一般形式中，公式(5)漂移项 $f_t(\boldsymbol{x})$不一定局限于$f(t)\boldsymbol{x}$，但在实际应用中通过此假设仍能保持高度的普适性，且对于重整流模型也有相似结论，因此讨论基于公式(8)是合理的。基于此，我们重新审视分数匹配模型的逆向SDE公式。假设满足 $\sigma_t = \sqrt{k}g_t$，我们可以对逆向过程进行如下数值近似分解：

$$\begin{aligned}
-\Delta\boldsymbol{x}_t &= \left(f_t(\boldsymbol{x}_t) - \frac{1+k}{2} g_t^2 \nabla_{\boldsymbol{x}_t} \log p_t\,(\boldsymbol{x}_t)\right)(-\Delta t) \\
&\quad + \sqrt{k}g_t\sqrt{\Delta t}\boldsymbol{\epsilon} \\
&= \left(f_t(\boldsymbol{x}_t) - \frac{1}{2} g_t^2 \nabla_{\boldsymbol{x}_t} \log p_t\,(\boldsymbol{x}_t)\right)[-(1+k)\Delta t] \\
&\quad + k f_t(\boldsymbol{x}_t)\Delta t + g_t\sqrt{k\Delta t}\boldsymbol{\epsilon} \\
&\approx \underbrace{\left(f_t(\boldsymbol{x}_t) - \frac{1}{2} g_t^2 \nabla_{\boldsymbol{x}_t} \log p_t\,(\boldsymbol{x}_t)\right)[-(1+k)\Delta t]}_{\text{ODE去噪过程}} \\
&\quad + \underbrace{f_{t-(1+k)\Delta t}\left(\boldsymbol{x}_{t-(1+k)\Delta t}\right)(k\Delta \mathrm{t}) + g_{t-(1+k)\Delta t}\sqrt{k\Delta t}\boldsymbol{\epsilon}}_{\text{正向加噪过程}},
\end{aligned}$$

其中$\boldsymbol{\epsilon} \sim \mathcal{N}(0,1)$。上述推导揭示了一个关键洞察：当$\sigma_t = \sqrt{k}g_t$时，一个随机逆向步骤在数值上等效于先对公式(5)执行一步ODE去噪（$\boldsymbol{x}_t \to \boldsymbol{x}_{t-\Delta t}$），随后执行一步前向加噪过程（$\boldsymbol{x}_{t-(1+k)\Delta t} \to \boldsymbol{x}_{t-\Delta t}$）。为了将这一发现推广，我们引入参数$p = 1/(1+k)$，并将时间步长重参数化，令$\Delta t \to \Delta t/(1+k)$。上述过程可抽象为路径$\boldsymbol{x}_t \to \boldsymbol{x}_{t-\Delta t} \to \boldsymbol{x}_{t-p\Delta t}$。在此过程中，边

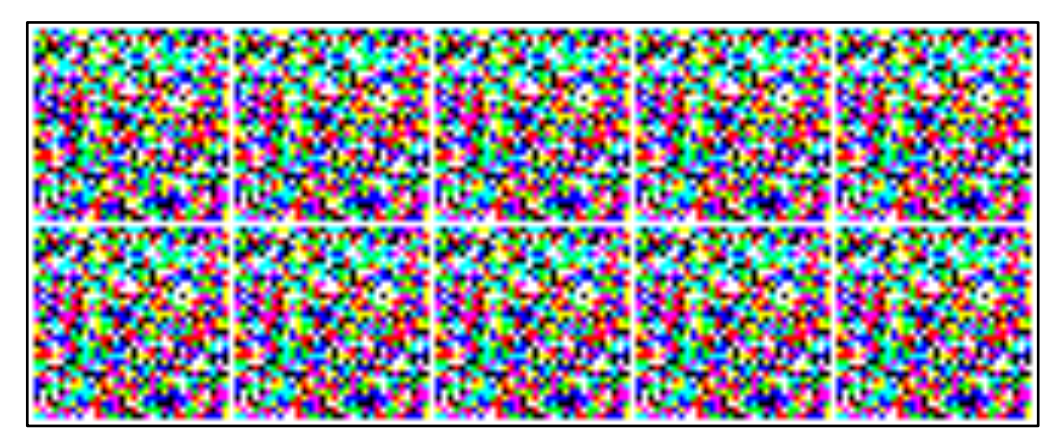

**图 2** 在分数匹配模型上，对模型输出直接分词的结果。从左到右，从上到下为同一个样本从$t=0$到$t=1$的样本。可以观察到，如果对模型输出直接做分词，则无法生成有意义的结果。

缘分布$p_t(\boldsymbol{x})$保持不变。当$p=0.5$时，该过程退化为标准的连续时间扩散模型。受此启发，我们在流匹配和一致性模型中引入类似的“去噪-再加噪”机制。尽管流匹配模型缺乏显式的中间态转换公式——如公式(6)，我们在3.1.4节中提出了一个统一形式——公式(22)——来解决此问题。至此，我们通过将前向过程$\boldsymbol{x}_{t-\Delta t} \rightarrow \boldsymbol{x}_{t-p\Delta t}$中添加的随机噪声$\boldsymbol{\epsilon}$进行离散化，便可以将DDCM的思想扩展至所有主流扩散变体。

参数$p$控制了前向加噪过程的回溯幅度。理论上，当$p=0.5$时，我们的方法在分数匹配模型上等价于连续版本的DDCM，在DDPM模型上等价于DDCM。虽然一般情况下应满足$p \in (0,1)$以构成合理的前向过程，但我们重点探讨$p=0$这一特殊情况。当$p=0$时，一次完整的“去噪-加噪”迭代（$\boldsymbol{x}_t \rightarrow \boldsymbol{x}_{t-\Delta t} \rightarrow \boldsymbol{x}_{t-p\Delta}$）使得样本的时间回归到初始时刻$t$。这种定点迭代策略具有参数空间解耦的优势：当$p \neq 0$时，分词效果与采样路径的超参数高度耦合。而$p=0$保持了边缘分布固定在$p_t(\boldsymbol{x})$，使得重建质量主要取决于当前时刻。定点策略允许在固定分布上反复优化编码，从而解耦了随机噪声与逆向轨迹质量之间的相互影响。在此模式下，我们需要在多次正逆向迭代结束后，执行常规的 ODE 逆向求解——公式(5)——以生成最终样本。

实验发现，不同类型的扩散模型对分词时刻$t$的敏感度存在显著差异：除了重整流模型之外的模型，都倾向于选择较小的$t$。这是因为在低噪声区域，条件分布满足$\nabla_{x_t} \log p_t(\boldsymbol{x}_t) \approx \nabla_{x_t} \log p_t(\boldsymbol{x}_t|\boldsymbol{x})$，能够有效保留$\boldsymbol{x}$的语义信息（如图 3、图 4所示）。重整流模型（Rectified Flow）： 倾向于选择较大的$t$。由于重整流模型经过了多次轨迹蒸馏（Reflow），其概率流速度场$\boldsymbol{v}(\boldsymbol{x}_t, t)$被拉直，使得即便在$t$较大时，速度场仍能精确指向目标数据，即$\boldsymbol{v}(\boldsymbol{x}_t, t) \approx \boldsymbol{v}(\boldsymbol{x}_t, t|\boldsymbol{x})$。这种全局一致性使得其在高噪区域仍能保持稳健的编码能力。

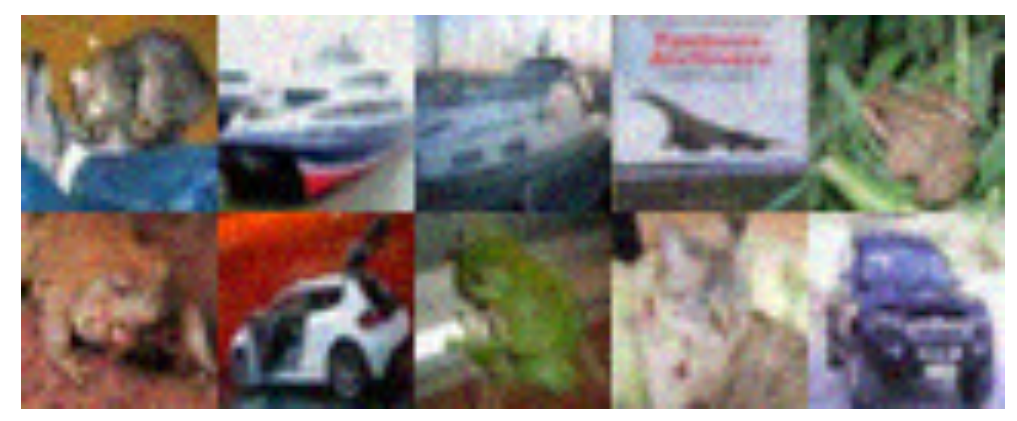

**图 3** 此图片展示了添加噪音较小的时候的样本，可以看出保留了原图片的大部分信息，模型可以几乎无损的重构原图。因此有$\nabla_{x_t} \log p_t(\boldsymbol{x}_t) \approx \nabla_{x_t} \log p_t(\boldsymbol{x}_t|\boldsymbol{x})$以及$(\boldsymbol{x}_t, t) \approx \boldsymbol{v}(\boldsymbol{x}_t, t|\boldsymbol{x})$。

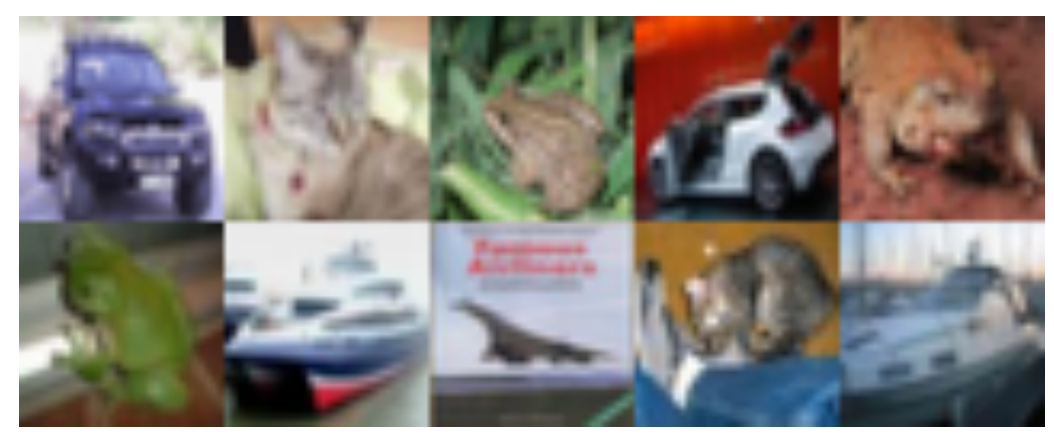

**图 4** 此图片展示的是图 3所示样本的重构图像，可以看出几乎能够重构原始图片。

在应用正逆向采样时，为了在有限计算预算下最大化编码效率，我们提出了一种由粗到精的自适应步长调度策略。首先，我们将采样步长$\Delta t$划分为$P$个区间。第$k$个区间的步长$\Delta t(k, \Delta t_{\min}, \Delta t_{\max})$满足：

$$\Delta t(k, \Delta t_{\min}, \Delta t_{\max}) = \Delta t_{\min}^{k/P} \Delta t_{\max}^{(P-k)/P}, \tag{9}$$

其次，对于总迭代次数$N$，我们动态分配每个区间的迭代次数$r(k, \sigma_{\max})$定义为：

$$r(k, \sigma_{\max}) = \left\lfloor \frac{(1 + k/P(\sigma_{\max}^{1/\rho} - 1))^{\rho}}{\sum_i (1 + i/P(\sigma_{\max}^{1/\rho} - 1))^{\rho}} N + 0.5 \right\rfloor. \tag{10}$$

公式(10)类似于在EDM[15]中提出的时间调度。

对于离散时间的去噪扩散模型（DDPM），我们利用 DDIM 的逆向过程进行适配。给定离散时间序列$\sigma_{\tau_{i-1}} < \sigma_{\tau_i} < \sigma_{\tau_{i+1}}$，更新公式如下：

$$x_{\tau_{i-1}} = \frac{\overline{\alpha}_{\tau_{i-1}}}{\overline{\alpha}_{\tau_i}} \left( x_{\tau_i} - \left( \overline{\beta}_{\tau_i} - \frac{\overline{\alpha}_{\tau_i}}{\overline{\alpha}_{\tau_{i-1}}} \sqrt{\overline{\beta}_{\tau_{i-1}}^2 - \overline{\sigma}_{\tau_i}^2} \right) \epsilon_\theta \right) + \overline{\sigma}_{\tau_i} \epsilon, \tag{11}$$

其中，$\overline{\sigma}_{\tau_i}$有：

$$\overline{\sigma}_{\tau_i} = p \frac{\overline{\beta}_{\tau_{i-1}}}{2\overline{\beta}_{\tau_i}} \sqrt{1 - \left( \frac{\overline{\alpha}_{\tau_i}}{\overline{\alpha}_{\tau_{i-1}}} \right)^2},$$

类似的，当$p=0.5$时，上述模型变为DDCM。

在$p=0$的离散化场景下，我们将公式(9)中的连续时间步长映射为离散步数索引$m_k$。对于某个时间步长$\Delta t_{\min} \in N$，令$\Delta t_{max} = \Delta t_{\min} + P$有：

$$\begin{aligned} \Delta t(k, \Delta t_{\min}, \Delta t_{\max}) &= \Delta t_{\min} + k \\ &= \Delta t_{\min}^{m_k/P} \Delta t_{\max}^{(P-m_k)/P}, \end{aligned} \tag{12}$$

从公式(12)解出$m_k$之后，带入公式(10)的离散形式得

到公式：

$$r(k)=\left\lfloor\frac{(1+m_k/P(\sigma_{\max}^{1/\rho}-1))^{\rho}}{\sum_i(1+m_i/P(\sigma_{\max}^{1/\rho}-1))^{\rho}}N+0.5\right\rfloor. \quad (13)$$

上述算法中的超参数均通过在验证集上进行低成本搜索确定。

#### 3.1.2 如何选择噪音

基于公式(8)的统一边缘分布，本节推导如何从码表中选择最优的噪声令牌$\boldsymbol{\epsilon}_c$以实现图像重构。假设当前时刻$t$的状态为$\boldsymbol{x}_t=s(t)\boldsymbol{x}+\Sigma(t)\boldsymbol{\epsilon}$，我们的目标是通过分词操作，使样本重构为目标图像$\boldsymbol{x}$。考虑应用一次$\boldsymbol{x}_t\rightarrow\boldsymbol{x}_{t-\Delta t}\rightarrow\boldsymbol{x}_{t-p\Delta t}$的“去噪-再加噪”过程。不失一般性，我们分析$p=0$的情形。在$t$较小的假设下（条件分布近似于边缘分布），更新后的样本$\boldsymbol{x}_{t-p\Delta t}$可近似表示为：

$$\boldsymbol{x}_{t-p\Delta t}=s(t)\boldsymbol{x}_0+\Sigma(t-\Delta t)\boldsymbol{\epsilon}+\Sigma(t)\sqrt{\Delta t}\boldsymbol{\epsilon}',$$

其中$\boldsymbol{\epsilon}'$是我们需要引入的编码噪声。为了使更新后的状态逼近目标分布，我们期望$\boldsymbol{x}_{t-p\Delta t}$等价于目标状态$\boldsymbol{x}_t'=s(t)\boldsymbol{x}+\Sigma(t)\boldsymbol{\epsilon}$。联立两式求解$\boldsymbol{\epsilon}'$，可得：

$$\boldsymbol{\epsilon}'=(s(t)(\boldsymbol{x}_0-\boldsymbol{x})+(\Sigma(t)-\Sigma(t-\Delta t))\boldsymbol{\epsilon})/(\Sigma(t)\sqrt{\Delta t})$$

我们的目标是从预定义的码表$\mathcal{E}_i$中选择一个噪声令牌$\boldsymbol{\epsilon}_t$，使其与理想噪声$\boldsymbol{\epsilon}'$的欧氏距离最小，即$\arg\min_{\boldsymbol{\epsilon}_c\in\mathcal{E}_i}\|\boldsymbol{\epsilon}_t-\boldsymbol{\epsilon}'\|^2$。展开目标函数，我们得到：

$$\begin{aligned}\|\boldsymbol{\epsilon}_t-\boldsymbol{\epsilon}'\|^2=&\|\boldsymbol{\epsilon}_c-(\Sigma(t)-\Sigma(t-\Delta t))\boldsymbol{\epsilon}/(\Sigma(t)\sqrt{\Delta t})\|^2\\&-2\boldsymbol{\epsilon}_t s(t)(\boldsymbol{x}_0-\boldsymbol{x})+\text{与}\boldsymbol{\epsilon}_t\text{无关项}\end{aligned}$$

在上述优化目标中，第一项描述了候选噪声$\boldsymbol{\epsilon}_t$与上一时刻残留噪声$\boldsymbol{\epsilon}$之间的相关性。由于码表中的噪声通常服从高斯分布且独立于历史噪声，因此该项可以被认为是固定值。忽略与$\boldsymbol{\epsilon}_t$无关的常数项及噪声干扰项，该最小化问题等价于最大化$\boldsymbol{\epsilon}_t$与残差方向的内积：

$$\arg\min_{\boldsymbol{\epsilon}_t\in\mathcal{E}_i}\|\boldsymbol{\epsilon}_t-\boldsymbol{\epsilon}'\|^2=\arg\max_{\boldsymbol{\epsilon}_t\in\mathcal{E}_i}\boldsymbol{\epsilon}_t^T(\boldsymbol{x}-\boldsymbol{x}_0), \quad (14)$$

至此，我们通过另一种方式给出了[12]中挑选噪音的规则。这一结论揭示了DDCM的本质： 噪声令牌的选择应当最大程度地与当前样本与目标样本的残差方向$\boldsymbol{x}-\boldsymbol{x}_0$保持一致。同时也解释了前文提到的现象：当$t$较大时，由于条件分布与边缘分布差异巨大，上述近似不再成立，导致编码效率显著下降，因此除了重整流模型，在$p=0$的时候，都需要选择$t$较小的时刻，才能取得较好的重构样本。这也暗示在DDCM中，重构过程从最大时间开始，可能造成时间步的浪费，如图 5所示。

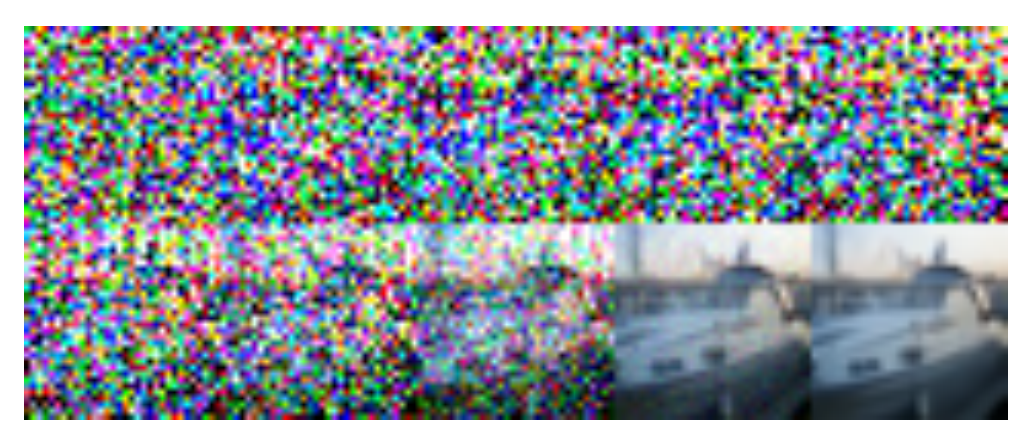

**图 5** 本图展示了DDCM的过程，可以发现算法在噪音较大的时刻执行多次。而此时包含信息较少，可能造成浪费。

值得注意的是，虽然我们的方法在每个时间步都需要使用不同的码表集合，但这并不构成存储瓶颈。利用伪随机数生成器的特性，我们无需物理存储所有码表向量。仅需保存码表的随机种子和索引，即可在推理过程中动态重建所需的“虚拟码表”。相比于模型推理的计算量，这种动态生成的开销几乎可以忽略不计，从而保证了方法的内存开销的高效性。

#### 3.1.3 初始点的选择

与从逆向过程的最大时刻进行迭代的DDCM不同，我们的方法从中间时刻$T_r$开始迭代（其中$T_r\leq T$）。在此时刻，数据的边缘分布$p_{T_r}(x)$不再满足标准正态分布假设（$p_{T_r}(x)\sim\mathcal{N}(0,1)$），而是包含了部分数据信息。为了探究不同初始化先验对分词及重建性能的影响，我们设计并评估了以下四种初始化策略：

1. 基于码表的初始化：利用公式(15)的度量标准，直接从码表中检索与当前数据分布最匹配的噪声令牌$\boldsymbol{\epsilon}_t$，随后将其缩放至$T_r$。这种策略试图在初始时刻即引入离散化的先验信息。
2. 高斯噪声近似：忽略$T_r$时刻的数据漂移项，仅保留随机性。从标准正态分布中采样噪声$\boldsymbol{\epsilon}\sim\mathcal{N}(0,1)$，随后将其缩放至$T_r$。
3. 利用扩散模型的边缘分布公式(8)，结合真实数据$\boldsymbol{x}\sim p_{\text{data}}$和采样噪声$\boldsymbol{\epsilon}\sim\mathcal{N}(0,1)$，构造理论上$T_r$时刻服从的数据分布。
4. 零初始化：采用确定性的全零张量作为起点。

$$\arg\max_{\boldsymbol{\epsilon}_t\in\mathcal{E}_i}\boldsymbol{\epsilon}_t^T\boldsymbol{x}. \quad (15)$$

我们在4.3.1节中展示了不同的初始化方式，重构效果的差异。

#### 3.1.4 统一形式

尽管不同变体的扩散模型采用不同的训练目标（如$\boldsymbol{\epsilon}$-prediction，$\boldsymbol{v}$-prediction或$\boldsymbol{x}$-prediction），但其核心输出均可被重参数化为样本$\boldsymbol{x}$与噪声$\boldsymbol{\epsilon}$的线性组

合。对于任意时刻$t$，设模型原始输出为$\boldsymbol{o}_t$。通用的训练目标通常最小化$\mathbb{E}\|\boldsymbol{o}_t-(a_t\boldsymbol{x}+b_t\boldsymbol{\epsilon})\|^2$。利用公式(8)中的线性关系，我们可以从模型输出中解出对$\boldsymbol{x}$的估计$\boldsymbol{x}_0$和$\boldsymbol{\epsilon}$：

$$\boldsymbol{o}_t = a_t\boldsymbol{x}+b_t\boldsymbol{\epsilon}, \tag{16}$$

公式(16)联立公式(8)即可求得预测的$\boldsymbol{x}_0$以及$\boldsymbol{\epsilon}$，且求得的$\boldsymbol{\epsilon}$等于损失函数为$\mathbb{E}\|\boldsymbol{o}_t-\boldsymbol{x}\|^2$的模型预测值。类似的，$\epsilon'$则等于损失函数为$\mathbb{E}\|\boldsymbol{o}_t-\boldsymbol{\epsilon}\|^2$的模型预测值。

对于流匹配模型来说，由于没有显式的正向过程，我们使用另一种方式来构建$\boldsymbol{x}_t\to\boldsymbol{x}_{t-\Delta t}\to\boldsymbol{x}_{t-p\Delta t}$的过程。对于分数匹配模型(5)来说，已有的结果有：

$$\nabla_x \log p_t(\boldsymbol{x}_t) = -(\boldsymbol{x}_t - s(t)\boldsymbol{x}_0)/\Sigma(t)^2, \tag{17}$$

其中$\Sigma(t)=s(t)\sigma(t)$。如上文所述，我们只讨论$f_t(\boldsymbol{x})$是线性函数的情况，令$f_t(\boldsymbol{x})=f_t\boldsymbol{x}$。则可以通过公式(18)获得对应时刻的样本：

$$x_{t-\Delta t} = s(t-\Delta t)\boldsymbol{x}_0 + \Sigma(t-\Delta t)\boldsymbol{\epsilon}. \tag{18}$$

**定理1** 公式(18)在$\mathcal{O}(\Delta t^2)$误差内，等价于使用欧拉公式对公式(5)进行一次迭代。

**证明**：为了方便书写，忽略二阶以上的高阶项，证明如下：

公式(5)使用欧拉法，有：

$$\boldsymbol{x}_{t-\Delta t} = \boldsymbol{x}_t - \left(f_t\boldsymbol{x}_t - \frac{1}{2}g_t^2\nabla_x \log p_t(\boldsymbol{x})\right)\Delta t$$

将公式(17)解出$\boldsymbol{x}_t$带入上式整理可以得到：

$$\begin{aligned}\boldsymbol{x}_{t-\Delta t} &= (s(t)-s(t)f_t\Delta t)\boldsymbol{x}_0 + \nabla_x\log p_t(\boldsymbol{x})\frac{1}{2}g_t^2\Delta t \\ &-\nabla_x\log p_t(\boldsymbol{x})\left(\left(s(t)\sigma(t)\right)^2-\left(s(t)\sigma(t)\right)^2\Delta t\right)\end{aligned} \tag{19}$$

根据分数匹配模型迭代的关系有$s(t-\Delta t)=s(t)-s(t)f_t\Delta t+\mathcal{O}(\Delta t^2)$，忽略高阶项，可得：

$$(s(t)-s(t)f_t\Delta t)\boldsymbol{x}_0 = s(t-\Delta t)\boldsymbol{x}_0$$

以及：

$$\left(s(t)\sigma(t)\right)^2-\left(s(t)\sigma(t)\right)^2\Delta t = s(t)s(t-\Delta t)\sigma(t)^2$$

带入公式(19)可得：

$$\begin{aligned}\boldsymbol{x}_{t-\Delta t} &= s(t-\Delta t)\boldsymbol{x}_0 - \nabla_x\log p_t(\boldsymbol{x})\times \\ &s(t)s(t-\Delta t)\left(\sigma(t)^2-\frac{g(t)^2}{s(t)s(t-\Delta t)}\frac{\Delta t}{2}\right)\end{aligned} \tag{20}$$

对于任意连续可导函数有$s(t-\Delta t)s(t+\Delta t)=s(t)^2+\mathcal{O}(\Delta t^2)$，令$t+(1/2)\Delta t$代替$t$则有：

$$s(t)s(t-\Delta t) = s(t+(1/2)\Delta t)^2$$

在分数匹配模型中有：

$$\sigma(t-\Delta t)^2 = \sigma(t)^2 - g(t)^2/s(t)^2\,\Delta t + \mathcal{O}(\Delta t^2)$$

带入到公式(20)中，则有：

$$\begin{aligned}\sigma(t)^2-\frac{g(t)^2}{s(t)s(t-\Delta t)}\frac{\Delta t}{2} &= \sigma\left(t+\frac{\Delta t}{2}\right)^2 \\ &= \sigma(t)\sigma(t-\Delta t)\end{aligned} \tag{21}$$

再将公式(17)以及公式(21)带入到公式(20)，则可得公式(18)。对于流匹配模型的线性形式也有相似的结论。■

而当模型是一致性模型的时候，因为$\boldsymbol{x}_0\sim p_{\text{data}}$，公式(18)变成了从$p_t(\boldsymbol{x}_t|\boldsymbol{x}_0)$中采样，此公式同样生成对应的边缘分布。

基于此，当构建$\boldsymbol{x}_t\to\boldsymbol{x}_{t-\Delta t}\to\boldsymbol{x}_{t-p\Delta t}$时，使用的公式为：

$$\begin{aligned}\boldsymbol{x}_{t-p\Delta t} &= s(t-p\Delta t)\boldsymbol{x}_0 + \Sigma(t-\Delta t)\boldsymbol{\epsilon} + \\ &\sqrt{\Sigma(t-p\Delta t)^2-\Sigma(t-\Delta t)^2}\boldsymbol{\epsilon}_c,\end{aligned} \tag{22}$$

其中，$\boldsymbol{\epsilon}_c$是从公式(14)得来。这种形式解决了在3.1.1中，没有扩展到流匹配模型上面的问题。

至此，除了离散形式的去噪扩散模型需要使用离散形式公式(11)外，其他所有连续形式的变体都可以应用公式(22)执行对图像的分词。我们在算法 1和算法 2中，使用伪代码的形式描述了我们的算法。

**算法 1**：广义去噪扩散模型（gDDCM，$p\neq 0$）

**输入**：边缘分布的参数$s(t)$和$\sigma(t)$；目标分词样本$\boldsymbol{x}_0$；训练好的，经过变换能够输出$\boldsymbol{x}_0$和$\boldsymbol{\epsilon}$的扩散模型$\theta(\boldsymbol{x}_t,t)$；总分词数量$N$；重构开始最大时间噪音时间$T_r$（可以跟模型训练时候的最大时间$T$不同），结束时间$T_t$；时间调度$w(k,T_r,T_t)$，输出为某一步的步长；初始化函数$I(\boldsymbol{x}_0,T)$；正常采样函数sample$(\boldsymbol{x}_t,t)$；

**输出**：噪音序列$L_{\epsilon_c}$，重构样本$\boldsymbol{x}_r$。

```
1:  t = T, x_t = I(x_0, T)
2:  for k in range(N):
3:      Δt = w(k, T_r, T_t)/p
4:      x_0, ϵ = θ(x_t, t)
5:      应用公式(14)获得ϵ_c以及对应索引i
6:      应用公式(22)获得x_{t-pΔt}
7:      t ← t - pΔt
8:      L_{ϵ_c}.append(i)
9:  x_r = sample(x_t, t)
10: return L_{ϵ_c}, x_r
```

# 4 实验

## 4.1 实验设置

### 4.1.1 模型配置

为了全面验证gDDCM的普适性，我们评估了2.1节所述的四种主流扩散模型及其变体：

**算法 2**：广义去噪扩散模型（gDDCM，$p=0$）

**输入**：边缘分布的参数$s(t)$和$\sigma(t)$；目标分词样本$\boldsymbol{x}_0$；训练好的，经过变换能够输出$\boldsymbol{x}_0$和$\boldsymbol{\epsilon}$的扩散模型$\theta(\boldsymbol{x}_t,t)$；总分词数量$N$；分词时间$T_s$；时间调度$w(k)$，输出为某一步的步长；初始化函数$I(\boldsymbol{x}_0,T_s)$；正常采样函数$\mathrm{sample}(\boldsymbol{x}_t,t)$；最大最小步长$\Delta t_{\min},\Delta t_{\max}$；参数$\sigma_{\max}$；不同步长区间$P$的数量。

**输出**：噪音序列$L_{\epsilon_c}$，重构样本$\boldsymbol{x}_r$

1: $t=T_s$，$\boldsymbol{x}_t=I(\boldsymbol{x}_0,T_s)$

2: **for** $k$ **in range**$(P)$:

3: $\quad\Delta t=\Delta t(k,\Delta t_{\min},\Delta t_{\max})$，公式(9)

4: $\quad$**for** _ **in range**$(r(k,\sigma_{\max}))$:

5: $\quad\quad\boldsymbol{x}_0,\boldsymbol{\epsilon}=\theta(\boldsymbol{x}_t,t)$

6: $\quad\quad$应用公式(22)获得$\boldsymbol{x}_{t-p\Delta t}$

7: $\quad\quad$应用公式(14)获得$\boldsymbol{\epsilon}_c$以及对应索引$i$

8: $\quad\quad t\leftarrow t-p\Delta t$

9: $\quad\quad L_{\epsilon_c}.\mathrm{append}(i)$

10: $\boldsymbol{x}_r=\mathrm{sample}(\boldsymbol{x}_t,t)$

11: **return** $L_{\epsilon_c}$，$\boldsymbol{x}_r$

去噪扩散概率模型①与分数匹配模型②（其中，分数匹配模型我们选择了EDM[15]）；一致性模型③；重整流模型 (Rectified Flow)④。对于重整流模型来说，该模型通过“重整（Reflow）”操作拉直采样轨迹。虽然多次重整能显著提升路径的线性度，但同时也使其概率流特征偏离了传统的扩散过程。为了验证我们方法在非标准扩散路径上的有效性，我们在CIFAR10上特别选取了3-rectified（3次蒸馏）的模型。而在LSUN Bedroom数据集上，受限于公开预训练模型的可用性，我们使用基础的1-rectified模型。

#### 4.1.2 数据集与超参数搜索

我们在以下两个标准数据集上进行了实验：CIFAR10（$32\times32$）：为了严谨地避免数据泄漏(Data Leakage)，我们严格在官方测试集包含的10,000张图像上进行评估。在超参数选择阶段，我们从测试集中随机抽取128张图像作为验证子集，通过网格搜索选取使LPIPS距离最小的参数组合。LSUN Bedroom ($256\times256$)：鉴于该数据集未提供官方测试集，我们使用其包含300张图像的验证集进行评估。同样的，我们从中随机抽取20张图像用于超参数搜索。

#### 4.1.3 基线对比

据我们所知，DDCM[12]是目前唯一利用预训练扩散模型进行图像分词的方法。因此，我们将DDCM设为主要的对比基线。

原文[12]中，DDCM仅适用于DDPM架构。为了在其他变体（如流匹配、一致性模型）上进行对比，我们将gDDCM在参数$p=0.5$时的设置视为DDCM的广义扩展版，并作为基线展示。我们在表 1和表2中报告了实验结果，并使用×表示DDCM不能直接应用到对应模型上。这不仅能直观对比DDCM与gDDCM的性能差异，也展示了原始DDCM策略在迁移至现代扩散架构时的局限性。

#### 4.1.4 评估指标

为了全面量化分词后的重建质量与生成图像的真实度，我们采用了五项广泛使用的评估指标，涵盖了从像素级保真度到高层语义一致性的多个维度：

分布与语义指标：我们使用FID[16]、LPIPS[18]和IS[17]指标从语义上来衡量生成样本分布与真实数据分布之间的距离及多样性；信号保真度：我们报告了SSIM[19]和PSNR[19]，用于量化图像在像素级别的绝对误差和结构保留程度。

在CIFAR10数据集上，我们报告所有上述五项指标。然而，对于LSUN Bedroom数据集，由于我们仅使用了包含300张图像的验证集进行测试，该样本量远低于计算FID和IS所需的统计稳定性阈值（通常需要10,000张以上样本）。因此，为了避免统计偏差导致的评估失真，我们在LSUN Bedroom数据集上仅报告LPIPS、SSIM和PSNR指标。

### 4.2 实验结果

我们在 CIFAR10 数据集上进行了全面评估，固定参数设置为：$N=300$，码表大小$K=1024$，不同步长数量$P=50$。表 1展示了不同模型在$p=0$与

① CIFAR10使用检查点为huggingface:：google/ddpm-cifar10-32, LSUN Bedroom使用检查点为huggingface：google/ddpm-bedroom-256

② CIFAR10使用的检查点为：https://nvlabs-fi-cdn.nvidia.com/edm/pretrained/edm-cifar10-32x32-cond-vp.pkl，LSUN Bedroom使用检查点为：https://openaipublic.blob.core.windows.net/consistency/edm_bedroom256_ema.pt

③ CIFAR10使用的检查点为：https://openaipublic.blob.core.windows.net/consistency/jcm_checkpoints/cd-l2/checkpoints/checkpoint_80，LSUN Bedroom使用检查点为：https://openaipublic.blob.core.windows.net/consistency/ct_bedroom256.pt

④ CIFAR10使用的检查点为：https://drive.google.com/file/d/12Mn5nfAwfz4Hcw7AifyhF2AZ4EtEXdu1/view，LSUN Bedroom使用检查点为：https://drive.google.com/file/d/1HUJ95q605YK4sH3yIysk8j3pNXALJi8t/view

$p = 0.5$下的性能对比。核心发现如下：1. 实验结果表明，在所有测试模型中，$p = 0$的设置均一致性地优于$p = 0.5$。特别是对于一致性模型，$p = 0.5$设置导致重构完全失败，而的$p = 0$策略则成功实现了有效分词与高质量重构。这有力地证明了gDDCM的有效性，并证明了gDDCM将DDCM其扩展为一个通用的分词框架。2. 在所有变体中，Reflow与EDM取得了最佳的综合性能。具体而言，ReFlow在LPIPS和FID上表现优异，而EDM在PSNR上略胜一筹。3. 在基线对比上，即使在原始DDCM的主场（DDPM模型）上，我们的$p = 0$策略 (FID 5.6) 也显著优于$p = 0.5$的基线结果(FID 7.7)，进一步验证了gDDCM的有效性。

LSUN Bedroom数据集定量结果表 2展示了分辨率为$256 \times 256$的LSUN Bedroom数据集上的评估结果。参数设置为$N = 500$，$K = 1024$，$P = 50$。与CIFAR10类似，gDDCM成功扩展到了所有选定的扩散模型变体。值得注意的是，在LSUN Bedroom数据集上，DDPM和EDM取得了最佳的分词效果，而ReFlow和一致性模型的表现相对较弱。这种性能翻转可能归因于底座模型的能力差异：在LSUN Bedroom任务上，我们使用的未蒸馏ReFlow和一致性模型的原生生成质量均弱于经过充分训练的DDPM和EDM。即便如此，我们的方法仍能适配这些模型并完成分词任务，且经过网格搜索优化的$p \neq 0$设置依然稳定优于$p = 0.5$的基线。

**表 1** gDDCM 在 CIFAR10 上面的表现。对于一致性模型，在$p = 0.5$时并没有成功重构样本，使用 × 表示。而在$p$列里面，每个模型第二行 × 代表原始 DDCM 不能够应用在对应扩散模型上，第一行 ✓ 则代表提出方法可以应用在当前模型上。当应用在 DDPM 模型上，且$p = 0.5$时，gDDCM 等价于对比的**基线方法 DDCM**。我们在$p$列使用 DDCM，且对应结果使用斜体表示。最好结果和第二好结果分别用粗体和下划线表示。

| 模型 | $p$ | FID↓ | LPIPS↓ | IS↑ | SSIM↑ | PSNR↑ |
|---|---|---|---|---|---|---|
| DDPM | 0 ✓ | 5.6 | 0.100 | 9.95 | 0.95 | 32.1 |
| | **DDCM** | *7.7* | *0.138* | *9.67* | *0.93* | *29.6* |
| EDM | 0 ✓ | **3.2** | 0.060 | 10.5 | **0.98** | 35.1 |
| | 0.5 × | 4.5 | 0.099 | 10.3 | 0.95 | 31.8 |
| CM | 0 ✓ | 4.3 | 0.078 | **10.9** | 0.96 | 33.3 |
| | 0.5 × | × | × | × | × | × |
| ReFlow | 0 ✓ | 4.3 | **0.049** | 10.1 | **0.98** | **36.2** |
| | 0.5 × | 21 | 0.190 | 8.80 | 0.84 | 23.2 |

**表 2** gDDCM 在 LSUN Bedroom 上面的表现。对于一致性模型，在$p = 0.5$时并没有成功重构样本，使用 × 表示。每个模型第二行 × 代表原始 DDCM 不能够应用在对应扩散模型上，第一行 ✓ 则代表提出方法可以应用在当前模型上。当应用在 DDPM 模型上，且$p = 0.5$时，gDDCM 等价于对比的**基线方法 DDCM**。我们在$p$列使用 DDCM，且对应结果使用斜体表示。我们用$p \neq 0$来表示在$p$所有取值中，对验证集网格搜索的最好值。最好结果和第二好结果分别用粗体和下划线表示。

| 模型 | $p$ | LPIPS↓ | SSIM↑ | PSNR↑ |
|---|---|---|---|---|
| DDPM | $\neq 0$ ✓ | 0.18 | 0.68 | 25.3 |
| | **DDCM** | *0.21* | *0.66* | *24.8* |
| EDM | $\neq 0$ ✓ | **0.17** | 0.69 | **25.5** |
| | 0.5 × | 0.23 | 0.63 | 24.0 |
| CM | 0 ✓ | 0.20 | 0.68 | **25.5** |
| | 0.5 × | × | × | × |
| ReFlow | $\neq 0$ ✓ | 0.20 | **0.77** | 25.1 |
| | 0.5 × | 0.21 | 0.76 | 24.8 |

### 4.3 消融实验

在本节中，我们对gDDCM的关键超参数及架构设计进行消融分析。为了提高实验效率，我们首先验证了样本规模的代表性：实验表明，在随机抽取的128张验证集图像上的评估结果与完整测试集高度一致。因此，本节所有实验均基于该128张图像的子集进行。除非另有说明，默认参数设置为：$p = 0$，步长个数$P = 50$，时间步数$N = 300$，码表大小$K = 1024$。

#### 4.3.1 初始化方式

我们对比了在3.1.3节中提出的四种初始化策略，结果如图 6所示。最佳策略为类型1——基于码表的投影初始化。相比于类型2——高斯噪声近似，该策略利用码表先验引入了确定性的结构信息，从而在起点处即降低了重建误差。值得注意的是，随机样本初始化表现最差。这是因为该策略虽然从真实数据分布$p_{\text{data}}$采样，但随机选取的样本在语义空间上与目标重构样本距离过远，导致模型难以在有限的步数内将其“拉回”至目标流形。基于此发现，除特别说明，本文所有实验均默认采用基于码表的投影初始化策略。

#### 4.3.2 码表大小

图 7展示了码表大小$K$对重构质量（LPIPS）的影响。实验观察到，随着$K$的增加，重构误差呈单调下降趋势。值得注意的是，在对数-对数坐标系下，LPIPS与码表大小呈现出近似线性的关系。这表明增

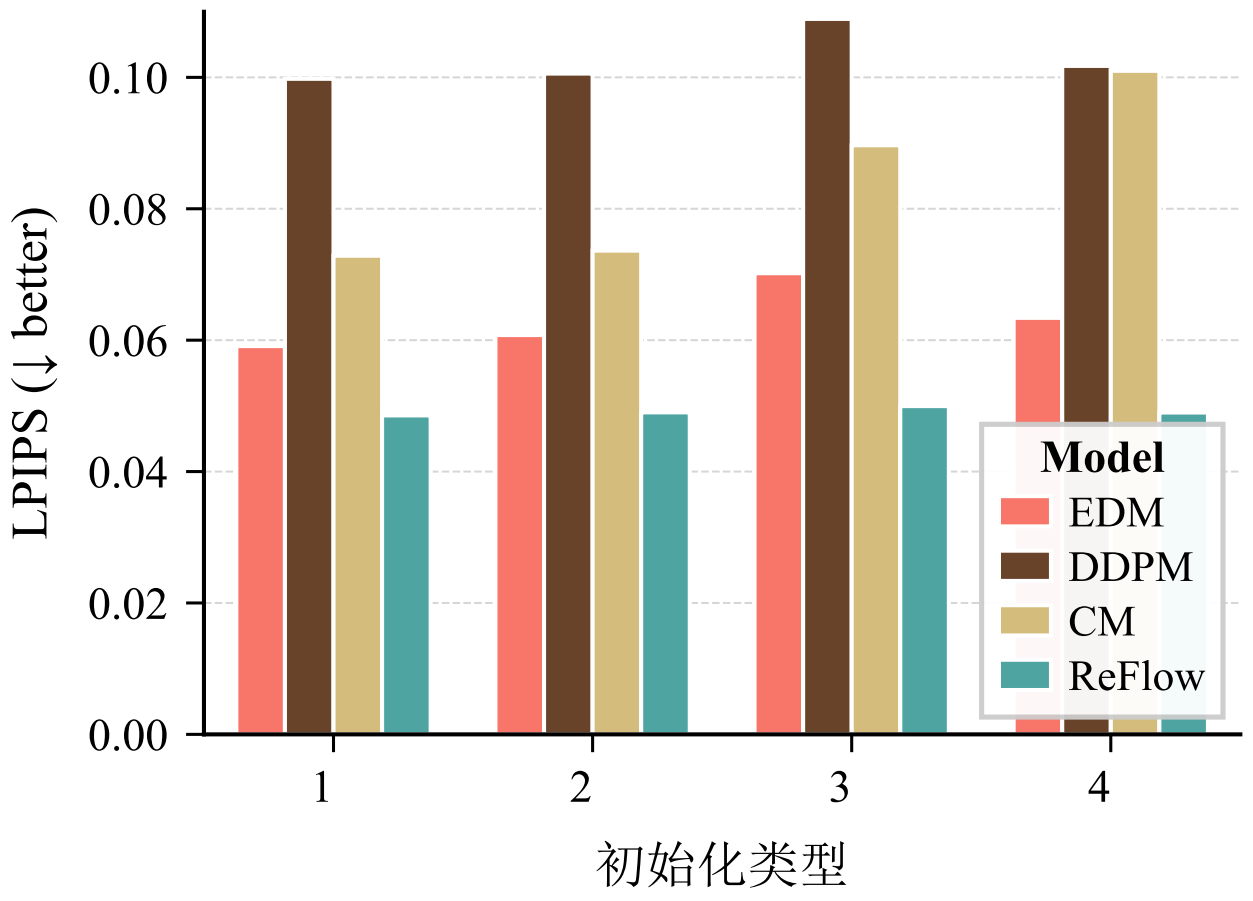

**图 6** 在CIFAR10上面不初始化的效果对比。横坐标分别对应于在3.1.3节中对应的四种方式。

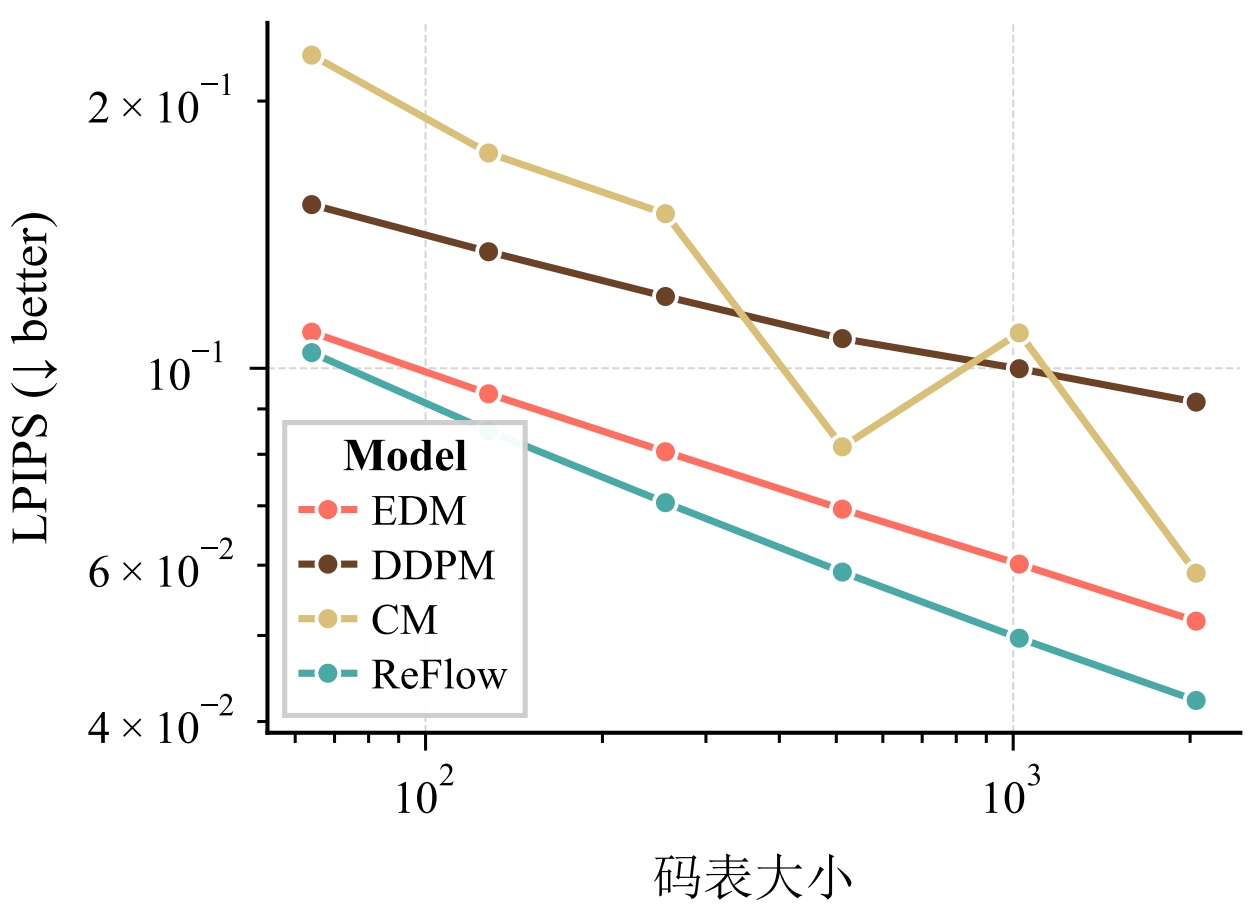

**图 7** 在CIFAR10上不同码表大小的效果对比。横纵坐标都为对数尺度。

加码表容量能有效提升重构能力，且在当前实验范围内尚未出现明显的性能饱和。

#### 4.3.3　分词数量

图 8展示了分词步数$N$（即生成的令牌序列长度）与重构质量的关系。总体而言，随着$N$的增加，绝大多数模型的重构损失显著降低。在所选模型中重整流模型表现出最快的下降速度，表明其平直的概率流轨迹更利于高效编码。

#### 4.3.4　2D分词

为了提升编码效率，我们也可以将gDDCM扩展至2D空间分词。具体而言，我们将图像平均分成多块（实验中取$2\times2$和$4\times4$），并对每个区块并行执行码表索引。在此设置下，总令牌数量变为$N\times$网格尺寸。关键优势在于，由于扩散模型的迭代次数仍由$N$决定，2D分词在不增加推理时间步数的情况下，显著提升了令牌的信息吞吐量（Throughput）。图 9报告了不同网格尺寸下的重构性能（固定$N=100$）。结果表明：我们的方法可以扩展到2D分词，并且更大的尺寸（如$4\times4$）显著提升了重构质量，其中ReFlow模型获益最大，表现远超其他模型。尽管2D分词具有低延迟优势，但空间网格引入了同一时间步内令牌的强相关性，这可能增加后续自回归模型建模的难度。因此，实际应用中需在重构质量、推理延迟与序列建模复杂度之间寻求平衡。

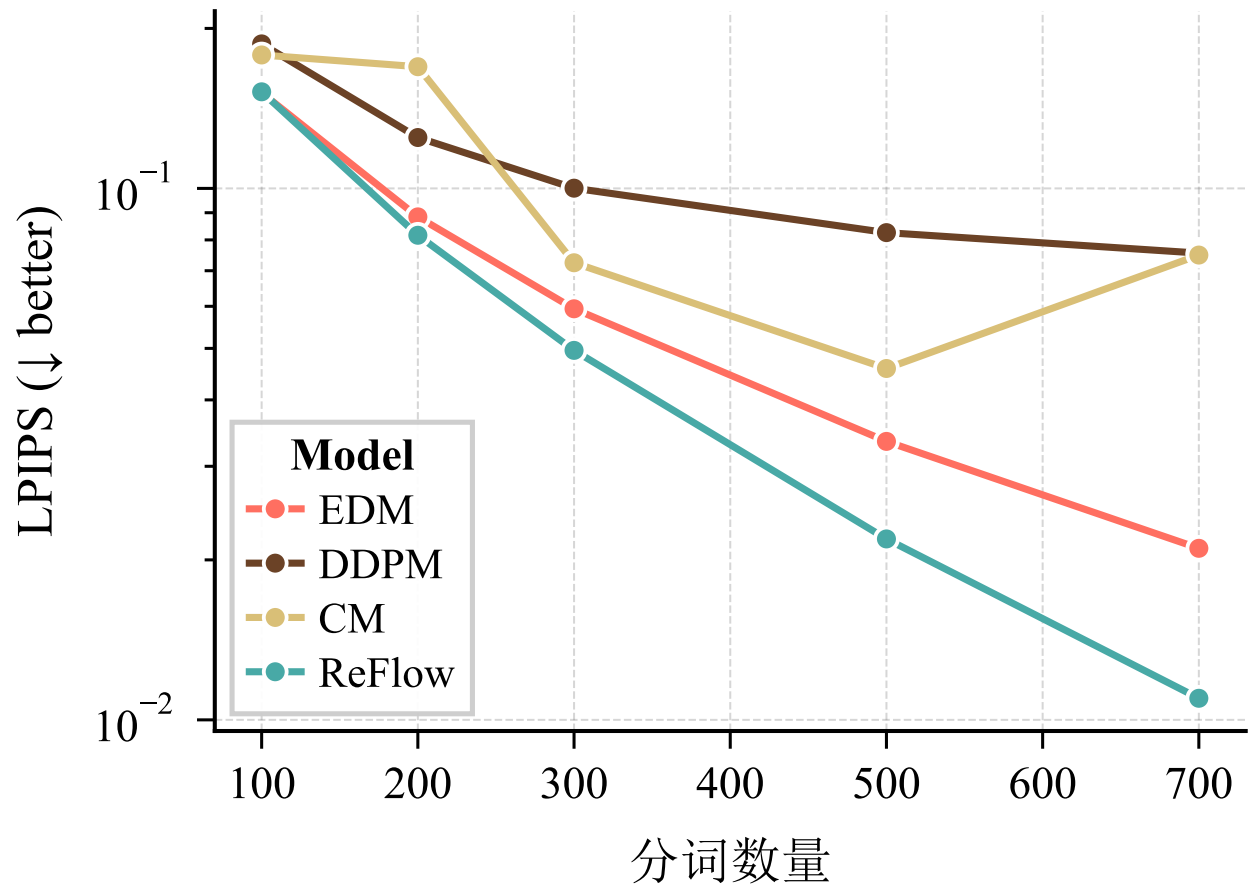

**图 8** 在CIFAR10上不同分词数量的效果对比。纵坐标为对数尺度。

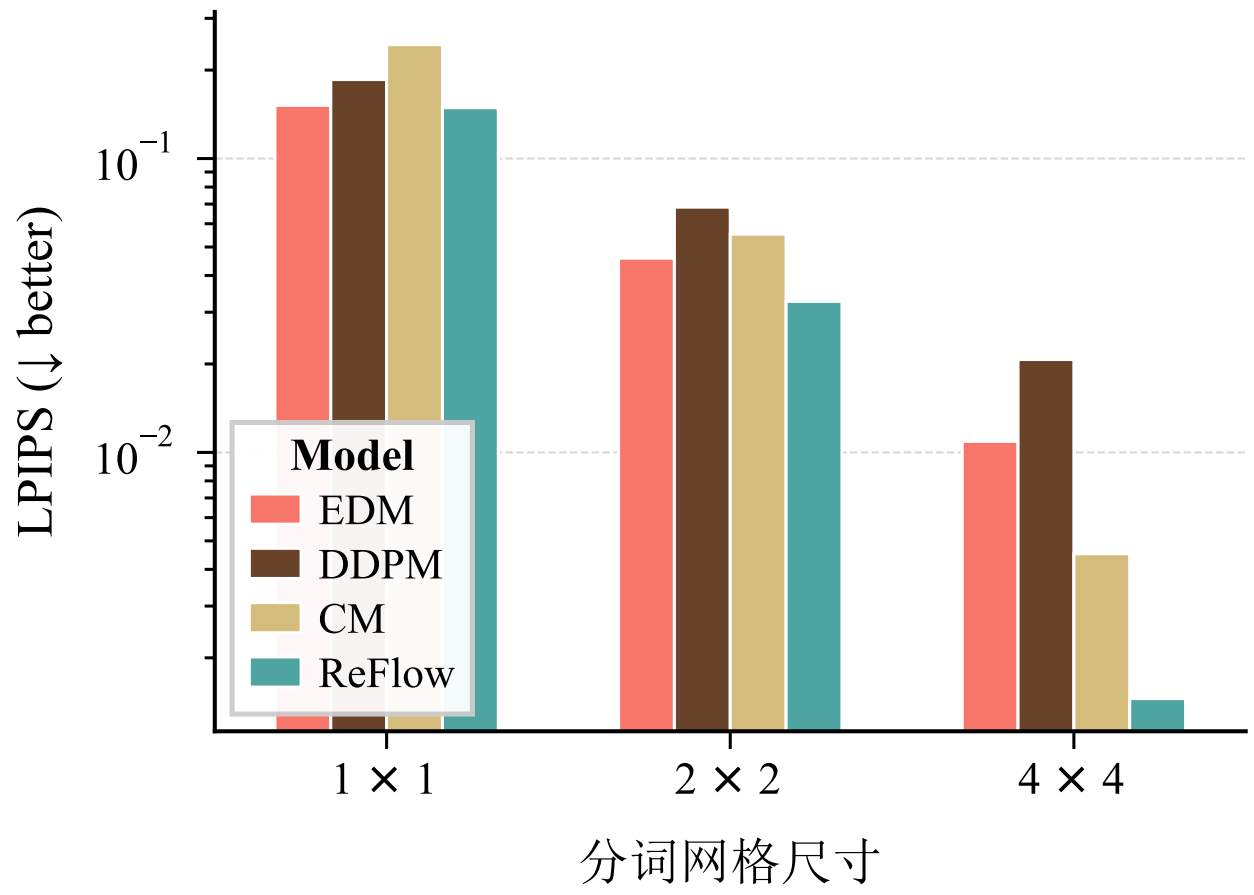

**图 9** 在CIFAR10上不同网格尺寸的效果对比。纵坐标为对数尺寸。

#### 4.3.5　$p$的取值影响

图 10展示了参数$p\in(0,1)$对不同模型性能的影响（$p=0$的结果详见表 1）。实验发现，对于所有测试模型，最优性能点均未出现在$p=0.5$处，而$p=0.5$对应于原始DDCM的设置。

这一结果表明：该设置在DDPM以及扩展至分数匹配、EDM或ReFlow等扩散模型时并非最优。这有力地验证了本文引入广义参数$p$的必要性，也证明了通过调整$p$值来适配不同模型采样轨迹的重要性。

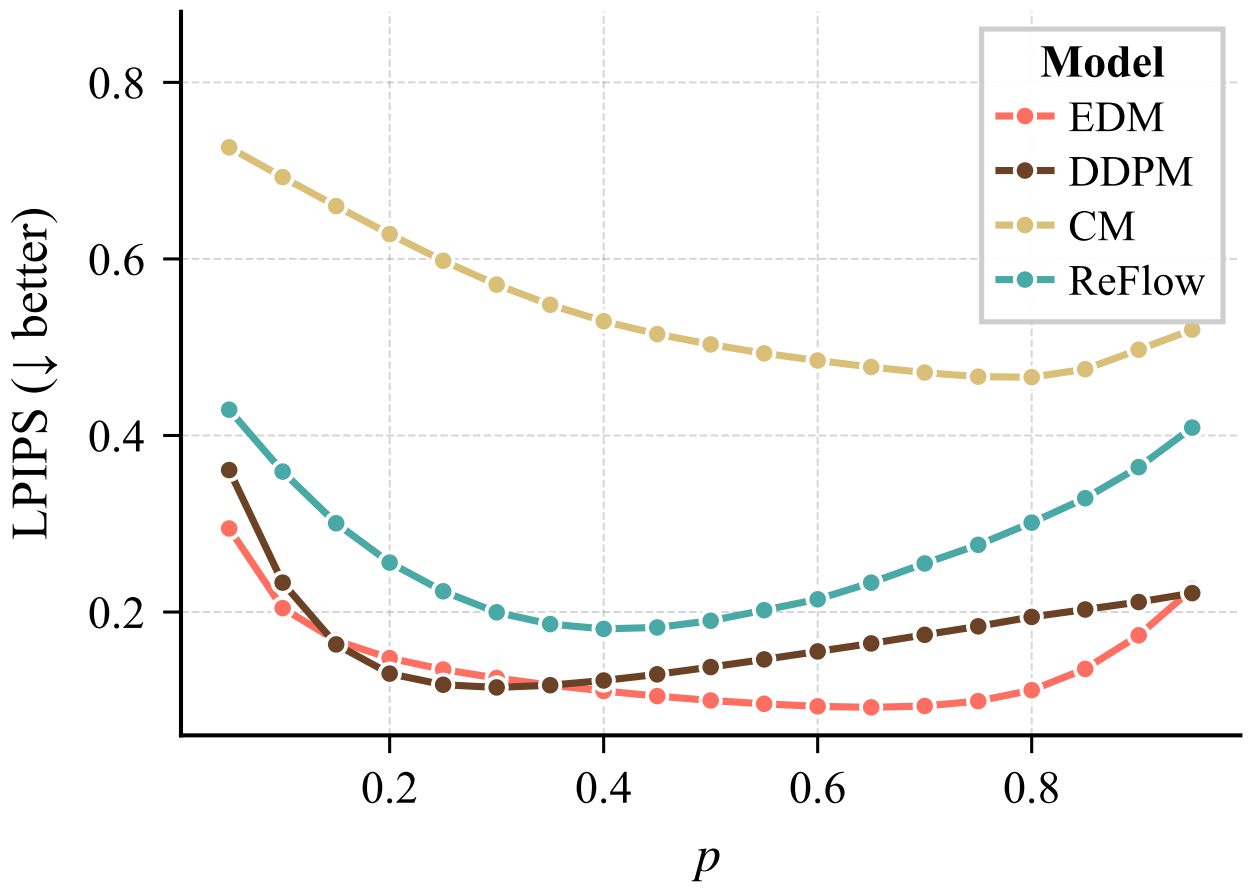

**图 10** 在CIFAR10上$p$取值对效果的影响。从$p = 0.05$每隔0.05采样至$p = 0.95$。

## 5 结论

本文深入剖析了去噪扩散编码模型（DDCM）在采样效率及架构兼容性方面的局限性。针对DDCM难以扩展至非DDPM模型（如流匹配模型）以及在高噪区域迭代效率低下的问题，我们提出了广义去噪扩散编码模型（Generalized Denoising Diffusion Coding Model, gDDCM）。本文的核心贡献在于构建了一个统一的分词理论框架。通过对扩散逆向过程的深度分析，我们提出了一种通用的“去噪-回溯（De-noise and Back-trace）”采样策略$\boldsymbol{x}_t \to \boldsymbol{x}_{t-\Delta t} \to \boldsymbol{x}_{t-p\Delta t}$。该策略巧妙地结合了基于常微分方程（ODE）的确定性去噪步骤与基于残差对齐的噪声注入步骤，并提出准对扩散模型的统一的迭代公式，不仅解决了难以扩展至非DDPM模型的问题，且显著提升了编码效率。基于此框架，我们成功将基于扩散的图像分词方法扩展至包括分数匹配模型、一致性模型及重整流模型在内的所有主流扩散模型变体。特别地，我们发现并验证了$p = 0$（即定点采样策略）在某些场景下的优越性，证明了完全解耦随机性与采样路径能够带来更稳定的分词效果。在CIFAR10和LSUN Bedroom数据集上的广泛实验表明，gDDCM在重构质量、感知保真度及模型普适性方面均全面优于现有的基线方法，且能够将1D采样分词扩展到2D。尽管gDDCM展现了卓越的性能，但其分词速度仍受限于扩散模型的推理步数。未来的工作将致力于探索如何结合蒸馏技术进一步减少所需的迭代次数，以及如何将生成的1D/2D令牌更高效地应用于大规模多模态生成模型的预训练中。

## 参 考 文 献

编 辑